\documentclass[conference]{IEEEtran}
\IEEEoverridecommandlockouts

\usepackage{cite}
\usepackage{amsmath,amssymb,amsfonts}
\usepackage{algorithmic}
\usepackage{graphicx}
\usepackage{multirow}
\usepackage{multicol}
\usepackage{textcomp}
\usepackage{xcolor}
\usepackage{placeins}
\usepackage{graphicx}
\usepackage{adjustbox}
\usepackage{booktabs}
\usepackage{colortbl}  

\def\BibTeX{{\rm B\kern-.05em{\sc i\kern-.025em b}\kern-.08em
    T\kern-.1667em\lower.7ex\hbox{E}\kern-.125emX}}

\begin{document}

\title{Towards Robust Federated Image Classification: An Empirical Study of Weight Selection Strategies in Manufacturing\\
}

\author{\IEEEauthorblockN{1\textsuperscript{st} Vinit Hegiste}
\IEEEauthorblockA{\textit{Chair of Machine Tools and Control Systems} \\
\textit{RPTU Kaiserslautern-Landau}\\
Kaiserslautern, Germany \\
0000-0001-6944-1988}
\and
\IEEEauthorblockN{2\textsuperscript{nd} Tatjana Legler}
\IEEEauthorblockA{\textit{Chair of Machine Tools and Control Systems} \\
\textit{RPTU Kaiserslautern-Landau}\\
Kaiserslautern, Germany \\
0000-0002-7297-0845}
\and
\IEEEauthorblockN{3\textsuperscript{rd} Martin Ruskowski}
\IEEEauthorblockA{\centerline{Innovative Factory Systems (IFS)} \\
\textit{German Research Center for Artificial Intelligence (DFKI)}\\
Kaiserslautern, Germany \\
0000-0002-6534-9057}
}

\maketitle

\begin{abstract}
In the realm of Federated Learning (FL), particularly within the manufacturing sector, the strategy for selecting client weights for server aggregation is pivotal for model performance. This study investigates the comparative effectiveness of two weight selection strategies: \textit{Final Epoch Weight Selection (FEWS)} and \textit{Optimal Epoch Weight Selection (OEWS)}. Designed for manufacturing contexts where collaboration typically involves a limited number of partners (two to four clients), our research focuses on federated image classification tasks. We employ various neural network architectures, including EfficientNet, ResNet, and VGG, to assess the impact of these weight selection strategies on model convergence and robustness.

Our research aims to determine whether FEWS or OEWS enhances the global FL model's performance across communication rounds (CRs). Through empirical analysis and rigorous experimentation, we seek to provide valuable insights for optimizing FL implementations in manufacturing, ensuring that collaborative efforts yield the most effective and reliable models with a limited number of participating clients. The findings from this study are expected to refine FL practices significantly in manufacturing, thereby enhancing the efficiency and performance of collaborative machine learning endeavors in this vital sector.
\end{abstract}

\begin{IEEEkeywords}
Federated Learning, Weight Selection Strategies, Manufacturing Collaboration, Federated Image Classification
\end{IEEEkeywords}

\section{Introduction}
Federated Learning (FL), a paradigm shift in distributed machine learning, enables multiple clients to collaboratively train a shared model while keeping their data localized, thus addressing critical issues of data privacy, security, and access rights \cite{mcmahan2017federated}. This novel learning approach is particularly relevant in sectors where data sensitivity and privacy are paramount, such as manufacturing, healthcare, and finance. In the manufacturing sector, FL opens avenues for leveraging distributed data sources across different facilities without compromising proprietary information or privacy \cite{smith2018federated, yang2019federated}. The potential applications of FL in manufacturing range from predictive maintenance to quality control and supply chain optimization, embodying a significant step towards intelligent manufacturing systems \cite{hegiste2022FedImg,hegiste2023FedOD}.

Despite the promising applications, much of the existing research on FL has focused on benchmark datasets like MNIST and CIFAR-10, often dividing these small datasets among a large number of clients to simulate distributed learning environments \cite{kairouz2021advances}. However, this approach is not reflective of real-world industrial settings. In manufacturing, the data generated is often redundant or repetitive, and the number of clients willing to participate in FL is typically small due to concerns about losing proprietary information, competitive advantage, or compliance with GDPR regulations. This research addresses these real-world constraints by focusing on a scenario with just 4 clients, where each client's dataset is unique but contains highly correlated and non-IID (non-Independent and Identically Distributed) data, which is common in industrial environments \cite{zhao2018federated}.
The FL setting required in such an environment is \textit{cross-silo} (e.g., FL between large institutions) \cite{kairouz2021advances}, where, due to the controlled environment and fewer clients, all clients are involved in each communication round (CR). However, the efficacy of a FL model largely hinges on the strategies employed for aggregating the updates from various clients. Algorithms such as FedProx \cite{li2020fedprox}, FedMA \cite{wang2020fedma}, FedNova \cite{wang2020Fednova} are various weight averaging algorithms, but they are built on the framework of the conventional FedSGD algorithm \cite{mcmahan2017federated}. The conventional FL framework aggregates weights based on the last epoch of local training, assuming it captures the most recent state of the model \cite{mcmahan2017federated}, which we refer to as \textit{Final Epoch Weight Selection} (FEWS) in this paper.

This research study provides insight into the comparative effectiveness of these weight selection strategies within a FL framework tailored for the manufacturing sector. Given the collaborative nature of manufacturing networks, often dominated by a limited number of partners, the choice of weight selection strategy could significantly influence the global model's performance. By focusing on federated image classification tasks, which is a critical component in manufacturing for defect detection, quality control, and process monitoring—this research aims to identify the optimal weight selection at the client side that promotes robust model convergence and efficacy. Through empirical analysis and experimentation, we seek to offer valuable insights for optimizing FL implementations in manufacturing, ensuring that collaborative efforts yield the most effective and reliable models with a limited number of participating clients. The findings of this study are poised to contribute significantly to the refinement of FL practices in manufacturing, enhancing the efficiency and performance of collaborative machine learning endeavors in this vital sector.

In this paper, we introduce and evaluate two weight selection strategies for FL: FEWS \cite{mcmahan2017federated, bharati2022fedapps}, and \textit{Optimal Epoch Weight Selection} (OEWS). FEWS involves sending the weights from the final epoch of local training to the server (which is the default setting), whereas OEWS strategy is introduced by us and involves selecting and sending the best-performing weights from the local training epochs. Our investigation centers on determining which strategy fosters a more robust and efficient global model, especially when testing the final global model on target objects in unseen environments, compared to the training environments.

\section{Related Work}

The concept of FL was first introduced by \cite{mcmahan2017federated}, who proposed a novel approach to machine learning in which a model is trained across multiple decentralized edge devices or servers holding local data samples without exchanging them. This foundational work laid the groundwork for subsequent research in the field, particularly in addressing challenges related to data privacy, efficiency, and scalability within FL frameworks.
In the manufacturing sector, \cite{smith2018federated} explored the application of FL for predictive maintenance in industrial IoT networks, highlighting the potential of FL to leverage distributed data sources while maintaining data privacy. Their work demonstrated the feasibility of FL in industrial settings, setting the stage for further exploration into more nuanced aspects such as model aggregation and weight selection strategies.
The selection of client weights for model aggregation in FL has been a significant area of interest, with various strategies proposed to enhance model performance. \cite{zhao2018federated} investigated the impact of non-IID data distributions on FL models and proposed strategies to mitigate the adverse effects through careful weight aggregation. Their findings underscore the importance of weight selection strategies in ensuring robust model performance in FL settings.

More recently, \cite{liu2023recentFLsurvey} conducted a comprehensive study on the efficacy of different weight selection strategies. Their empirical analysis revealed significant differences in model convergence and robustness depending on the weight aggregating strategy employed, highlighting the need for further research in this area, particularly in domain-specific contexts such as manufacturing. This paper extends that research by introducing our own OEWS strategy and comparing it with the default FEWS strategy based on a real industrial manufacturing use case.
Several studies have explored the application of FL across various sectors, demonstrating the promising growth of FL research over the years. \cite{bharati2022fedapps} provided a comprehensive overview of FL, its challenges, and its applications, underscoring its relevance across different industries. 
\cite{zhou2024MRI} leveraged EfficientNet-B0 and the FedAvg algorithm to explore the application of FL in medical imaging, specifically for MRI brain tumor detection, demonstrating how FL can facilitate collaboration among hospitals without sharing sensitive patient data. Recent work by \cite{hegiste2024enhancing} has shown that FL (FL) trained with clients having only real and synthetic data respectively not only outperforms centralized models trained on hybrid datasets but also benefits significantly from the inclusion of synthetic datasets from different clients, resulting in robust global models with enhanced generalization across diverse environments.
\cite{li2023cropdisease} evaluated convolutional network models and Vision Transformers using FL for image classification (vision) based crop disease detection. \cite{hegiste2022FedImg}, \cite{hegiste2023FedEnsemble}, and \cite{hegiste2023FedOD} have contributed to the literature by exploring quality inspection use cases in manufacturing via federated image classification and object detection, showcasing the advantages of FL in this domain. Furthermore, \cite{legler2024federatedcluster} explore the use of clustering techniques in FL systems to address challenges related to data heterogeneity and computational variability, proposing strategies for optimizing client selection and improving system efficiency and stability.

This highlights the recent growth in application-based FL research. However, there is a noticeable gap in survey papers or algorithm introduction papers that comprehensively compare multiple deep learning architectures for specific custom use cases using a FL approach. These comparisons are crucial for identifying efficient and robust architectures among the popular deep learning models, particularly when tailored to specific industrial applications.

\section{Methodology} \label{methodology}

This research examines the comparative effectiveness of Final Epoch Weight Selection (FEWS) versus Optimal Epoch Weight Selection (OEWS) in FL frameworks within the manufacturing sector, where client participation is limited. All models leverage pre-trained ImageNet weights to ensure faster convergence and to simulate real-world scenarios. Preliminary experiments using an ImageNet subset confirmed that pre-trained weights can achieve target accuracy in minimal epochs and communication rounds. Consequently, this study focuses on a custom dataset involving 3D-printed cabins with different windshield types.

\begin{table}[h]
    \centering
    \caption{Label Classification for Different Clients}
    \begin{tabular}{|c|c|c|c|c|c|}
    \hline
    \textbf{Client} & \textbf{No\_windshield} & \textbf{Type A} & \textbf{Type B} & \textbf{Type C} & \textbf{Type D} \\ \hline
    Client 1         & \checkmark                       & -               & \checkmark               & \checkmark               & \checkmark               \\ \hline
    Client 2         & \checkmark                       & \checkmark               & \checkmark               & \checkmark               & -               \\ \hline
    Client 3         & \checkmark                       & \checkmark               & \checkmark               & -               & \checkmark               \\ \hline
    Client 4         & \checkmark                       & \checkmark               & -               & \checkmark               & \checkmark               \\ \hline
    \end{tabular}
    \label{tab:windshield_types}
\end{table}
\begin{figure*}[h]
\centering
    \includegraphics[width=\textwidth]{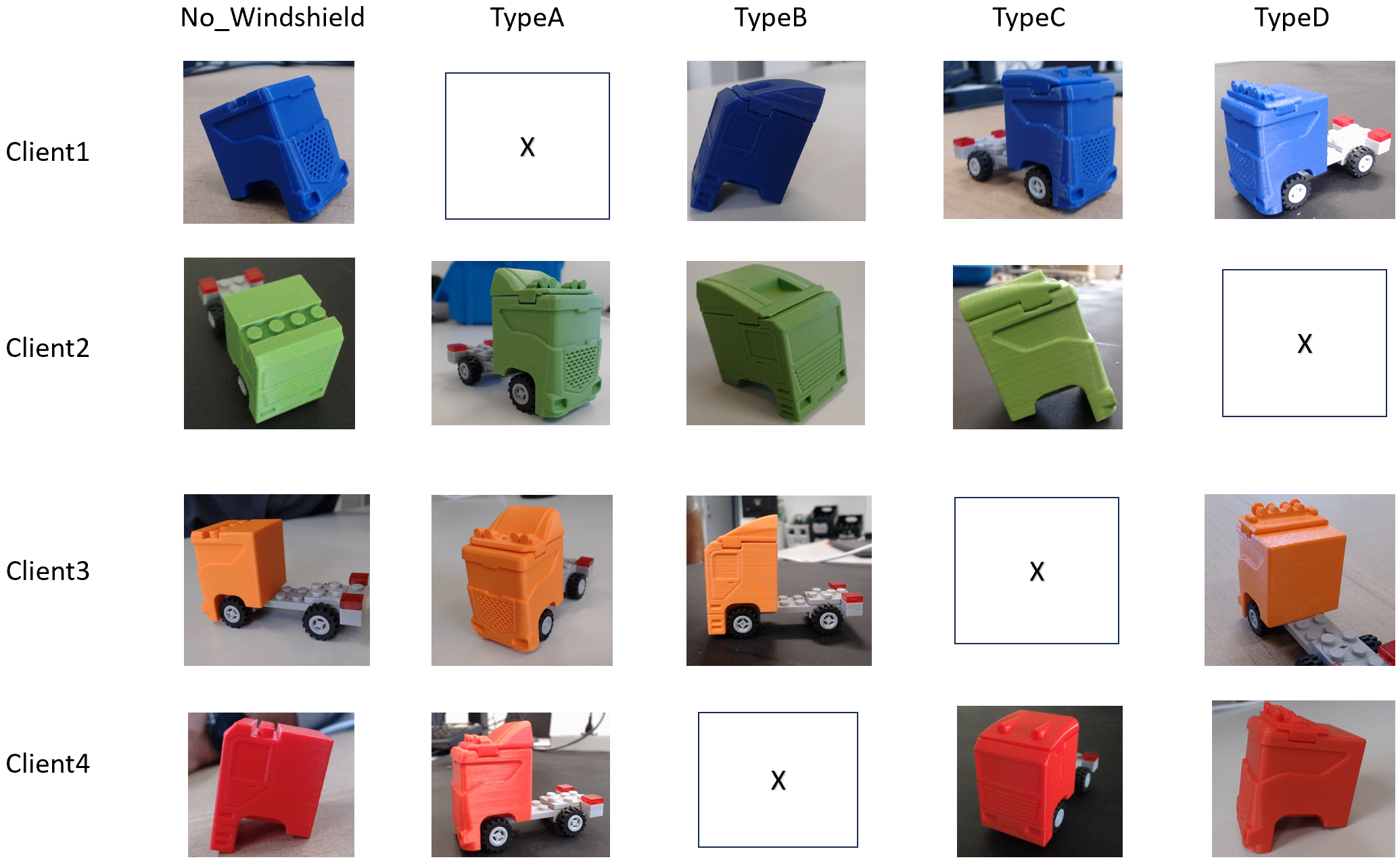}
    \caption{Visualization of class distribution and sample data for each client in the federated learning process.}
    \label{fig:training}
\end{figure*}
\begin{table*}[h]
    \centering
    \caption{Distribution of Images per Class Across Training, Validation, and Test Sets for Each Client}
    \begin{adjustbox}{max width=\textwidth}
    \begin{tabular}{ccccccccccccc}
        \toprule
        \textbf{Client} & \textbf{Set} & \textbf{No\_windshield} & \textbf{Windshield\_TypeA} & \textbf{Windshield\_TypeB} & \textbf{Windshield\_TypeC} & \textbf{Windshield\_TypeD} \\ \midrule
        \multirow{3}{*}{Client1} & Train & 409 & \cellcolor{red!50}0 & 404 & 440 & 441 \\ 
        & Val & 87 & \cellcolor{red!50}0 & 86 & 94 & 94 \\ 
        & Test & 89 & 100 & 88 & 95 & 96 \\ \midrule
        
        \multirow{3}{*}{Client2} & Train & 382 & 398 & 375 & 394 & \cellcolor{red!50}0 \\ 
        & Val & 82 & 85 & 80 & 84 & \cellcolor{red!50}0 \\ 
        & Test & 83 & 86 & 81 & 85 & 77 \\ \midrule
        
        \multirow{3}{*}{Client3} & Train & 408 & 427 & 445 & \cellcolor{red!50}0 & 422 \\ 
        & Val & 87 & 91 & 95 & \cellcolor{red!50}0 & 90 \\ 
        & Test & 89 & 93 & 97 & 86 & 91 \\ \midrule
        
        \multirow{3}{*}{Client4} & Train & 362 & 352 & \cellcolor{red!50}0 & 360 & 352 \\ 
        & Val & 77 & 75 & \cellcolor{red!50}0 & 77 & 75 \\ 
        & Test & 79 & 76 & 77 & 78 & 76 \\ 
        \bottomrule
    \end{tabular}
    \end{adjustbox}
    \label{tab:client_samples}
\end{table*}
\begin{figure}[h]
\centering
    \includegraphics[width=0.45\textwidth]{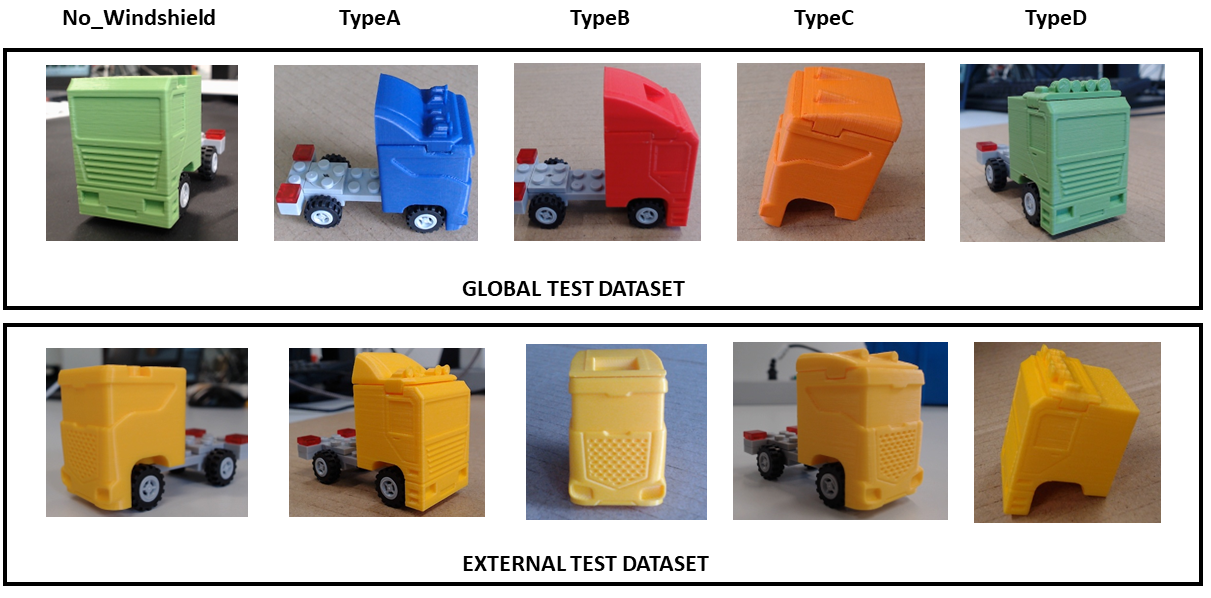}
    \caption{Test Datasets for Evaluating Federated Global Models, Clients' Local Centralized Models, and Entire Dataset Centralized Models.}
    \label{fig:test_global_and _external}
\end{figure}
\begin{table*}[h]
    \centering
    \caption{Distribution of Images per Class in the Global Test Set and External Client Test Set}
    \begin{adjustbox}{max width=\textwidth}
    \begin{tabular}{cccccc}
        \toprule
        \textbf{Dataset} & \textbf{Windshield\_TypeD} & \textbf{Windshield\_TypeA} & \textbf{Windshield\_TypeB} & \textbf{Windshield\_TypeC} & \textbf{No\_windshield} \\ \midrule
        \textbf{Global Test Set} & 340 & 355 & 343 & 344 & 340 \\ 
        \textbf{External Client Test Set} & 96 & 100 & 88 & 95 & 89 \\ 
        \bottomrule
    \end{tabular}
    \end{adjustbox}
    \label{tab:test_datasets}
\end{table*}
\subsection{Dataset} \label{dataset}

The application involves various 3D-printed cabin manufacturers, each producing different windshield types for cabins (Figure \ref{fig:training}). The classification labels for each client's dataset are shown in Table \ref{tab:windshield_types}. Each client lacks one class of images, simulating a scenario where a client intends to add the missing windshield type later in the manufacturing process. For experimental purposes, a global test set was created by combining test images of all clients' cabins, including classes that were missing during training. This setup evaluates the model's performance on local client datasets when encountering different cabin colors, windshields, or unseen windshield types.

To further assess FL models, an external dataset is introduced. This dataset, containing cabins with different colors and types of windshields (Figure \ref{fig:test_global_and _external}), is crucial for evaluating model generalization to unseen data. Table \ref{tab:test_datasets} presents the number of samples in both test sets used in this experiment.

\subsection{Federated Learning Process and Architecture} \label{architecture}

The Federated Learning (FL) process was implemented using the FLOWER framework \cite{beutel2020flower}, incorporating a custom plain averaging strategy. Unlike FLOWER's default weighted averaging, plain averaging simply averages model weights across clients. Previous research has shown that plain averaging can lead to faster convergence and improved accuracy, particularly in cases where client datasets are highly correlated \cite{hegiste2024collaborative}.

\begin{figure}[h]
\centering
    \includegraphics[width=0.35\textwidth]{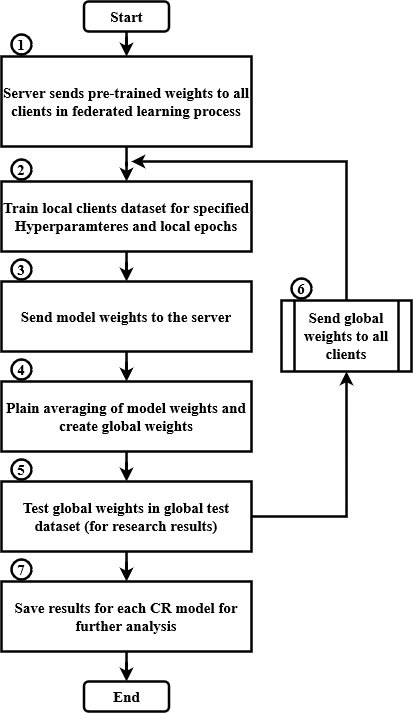}
    \caption{Proposed flow diagram for the training and evaluation process in a research-focused federated learning deployment within academic settings.}
    \label{fig:flow_flta}
\end{figure}

\begin{figure}[h]
\centering
    \includegraphics[width=0.35\textwidth]{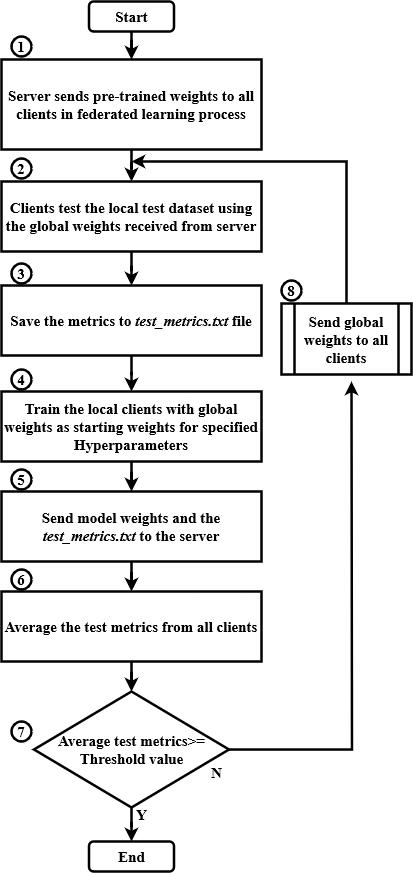}
    \caption{Proposed flow diagram for the training and evaluation process in real-world federated learning deployment within industrial settings.}
    \label{fig:flow_flta_deploy}
\end{figure}

In this study, all clients participated in each communication round (CR) during training. The academic FL process, illustrated in Figure \ref{fig:flow_flta}, follows the traditional FL process from steps 1 to 4, with additional steps such as testing global weights on a global test dataset (Step 5) and saving metrics to a \texttt{.txt} file (Step 7) for further analysis. These steps were conducted to identify the optimal global weights and halt training once a desired accuracy or F1 score was achieved, or to continue the FL process for further analysis. However, in practical industrial FL deployments, halting training to identify the best global model using a global test dataset is impractical, as it contradicts the core FL principle of non-sharing raw training data between clients and the server.

\textbf{For industrial applications, we propose an alternative workflow depicted in Figure \ref{fig:flow_flta_deploy}}. In this process, once the global model is received, it is evaluated by each client on their local test datasets (Step 2). These evaluations allow each client to assess the global model’s effectiveness within its specific environment before proceeding with further local training. This approach ensures that performance metrics accurately reflect the global model's effectiveness across diverse client environments, making it more suitable for industrial settings.
After evaluation, each client shares their test performance metrics and updated model weights with the server (Step 5). These metrics are aggregated by the server across all clients, and the plain averaging strategy is applied to update the global model weights (Step 6). Training is halted once the aggregated metrics surpass a predefined threshold, indicating that the global model has reached optimal performance (Step 7). This method is particularly relevant in environments where all clients contribute to each CR, ensuring that the global model remains effective and relevant.
In addition to the FL models, we also trained a centralized model by combining the training and validation data from all clients. This centralized model serves as a baseline for comparing the performance of FL models with traditional deep learning approaches. By comparing these models, we can evaluate the effectiveness of FL in preserving data privacy while achieving comparable or superior performance.

For all experiments, we standardized the FL configuration across all architectures, using a learning rate of 0.0001, a batch size of 16, and Stochastic Gradient Descent (SGD) with momentum (0.9) as the optimizer. The effectiveness of SGD with momentum is supported by studies such as \cite{wang2020Fednova} and \cite{wang2020slowmo}, which demonstrate a 3\% to 5\% performance improvement over standard SGD. The weight aggregation strategy employed was FedAvg \cite{mcmahan2017federated}, based on plain averaging.
This research evaluates the performance of four deep learning architectures: VGG19 \cite{simonyan2014vgg}, ResNet50 \cite{he2016resnet}, DenseNet121 \cite{huang2017densenet}, and EfficientNetv2 \cite{tan2021efficientnetv2}, under both centralized and FL setups. Centralized models were trained on individual clients' datasets and tested on both a global test dataset and an external client dataset. This analysis is particularly relevant in manufacturing environments, as it highlights the performance differences between centralized and FL approaches. The selection of deep learning architectures was influenced by factors such as the number of trainable parameters, ease of deployment, and availability of pre-trained weights. The primary goals of this research are to determine the optimal weight selection strategy between FEWS and OEWS and to identify the most suitable architecture for industrial deployment, considering factors such as model trainable parameters, performance in unseen environments, and overall feasibility for real-world applications.

\subsection{Plain Averaging with Optimal Epoch Weight Selection}\label{plainavg_oews}

In our FL framework, we enhanced the plain averaging strategy by: (1) selecting the best-performing weights from a fixed number of local epochs and (2) halting the training process once a desired threshold is reached. The updated global model weights \( w_{t+1} \) at round \( t+1 \) are calculated as:

\[
w_{t+1} = \frac{1}{K} \sum_{k=1}^K w_{t+1}^{k*}
\]

where:
\begin{itemize}
    \item \( w_{t+1} \) represents the global model weights after round \( t+1 \).
    \item \( w_{t+1}^{k*} \) denotes the best-performing model weights from client \( k \) within a fixed number of local epochs.
    \item \( K \) is the total number of clients participating in the FL process.
\end{itemize}

Training is halted when the aggregated global model achieves a predefined performance threshold \( T \) (e.g., accuracy, F1 score):

\[
\text{Stop training if: } \mathcal{M}(w_{t+1}) \geq T
\]

where:
\begin{itemize}
    \item \( \mathcal{M}(w_{t+1}) \) represents the evaluation metric of the global model after round \( t+1 \).
    \item \( T \) is the predefined performance threshold.
\end{itemize}

This method ensures that the global model is updated using the best local weights while preventing unnecessary communication rounds once satisfactory model performance is achieved.

\subsection{Experiment Setup} \label{experimental_setup}

The primary objective of this research is to evaluate FL weight selection strategies by comparing the default \textit{Final Epoch Weight Selection (FEWS)} with the proposed \textit{Optimal Epoch Weight Selection (OEWS)}. In the standard FL setup, client weights are sent to the server after a fixed number of local epochs, as per the FEWS strategy. The OEWS strategy, however, involves selecting and transmitting the best-performing local weights from each training round (refer to Section \ref{plainavg_oews}).
A secondary objective is to identify the most suitable deep learning architecture for FL, particularly when applied to our custom dataset. We evaluated four architectures under both FEWS and OEWS strategies.

To evaluate these strategies, two test datasets were used (Figure \ref{fig:test_global_and _external}, Table \ref{tab:test_datasets}):
\begin{itemize}
    \item \textbf{Global Test Dataset:} Combines test sets from all clients, including classes absent during training, assessing overall performance and generalization.
    \item \textbf{External Client Test Dataset:} Contains cabins with all five classes in different colors, simulating a new client or product variant to test the model's robustness to unseen data.
\end{itemize}
Centralized models were trained on a combined dataset from all clients’ training and validation data (Figure \ref{fig:training}). These models were trained for 100 epochs with early stopping at 30 epochs, matching the total effective epochs used in FL (15 local epochs × 5 CRs). We compared these centralized models with the federated models using both test datasets, focusing on generalization capabilities.
This setup provides a comprehensive evaluation of FL's effectiveness in manufacturing environments and aims to identify the most suitable deep learning architecture for FL with non-IID data.

\section{Experimental Results and Discussion} \label{Results}

This section presents the results based on the experimental setup described in Section \ref{experimental_setup}, focusing on outcomes from training clients with the datasets illustrated in Figure \ref{fig:training} and the classes detailed in Table \ref{tab:windshield_types}. The global models were trained using the federated architecture shown in Figure \ref{fig:flow_flta}, with testing conducted on the global test dataset (Figure \ref{fig:test_global_and _external}, Table \ref{tab:test_datasets}) after each communication round (CR). Additionally, we gathered results for the best-performing weights obtained via early stopping from centralized models trained on local clients' datasets and the entire centralized dataset. Various configurations of local epochs and CRs were tested, with two setups yielding the best results on both the global and external datasets: 10 local epochs with 10 CRs and 15 local epochs with 5 CRs. We proceeded with 15 epochs and 5 CRs, given the practical advantage of fewer communication rounds, which reduces the overhead of transferring weights to the server. Therefore, the results for FEWS and OEWS are based on this configuration.

\subsection{Comparison of Optimal Epoch Weight Selection (OEWS) vs. Final Epoch Weight Selection (FEWS) on Global Test Dataset}

The evaluation was conducted using two test datasets: a global test dataset (a combination of the test datasets from all clients) and an external client dataset, which differs from the training data. Experiments were conducted over 5 CRs, recording test accuracy, precision, F1 score, and recall for each CR. But in the results we mention the metrics from the final 5\textsuperscript{th} CR.

\begin{table}[h]
    \centering
    \caption{Comparison of performance metrics for various models and federated strategies on the global test dataset, highlighting the superior performance of Optimal Epoch Weight Selection (OEWS) over Final Epoch Weight Selection (FEWS).}
    \begin{adjustbox}{max width=\columnwidth}
    \begin{tabular}{ccccccc}
        \toprule
        \textbf{Model} & \textbf{Strategy} & \textbf{Test Acc} & \textbf{Precision} & \textbf{Recall} & \textbf{F1} \\ 
        \midrule
        \multirow{2}{*}{ResNet} 
        & FEWS & 0.9745 & 0.9748 & 0.9747 & 0.9745 \\ 
        & OEWS & \textbf{0.9890} & \textbf{0.9890} & \textbf{0.9890} & \textbf{0.9890} \\ 
        \midrule
        \multirow{2}{*}{VGG} 
        & FEWS & 0.9669 & 0.9671 & 0.9672 & 0.9669 \\ 
        & OEWS & \textbf{0.9669} & \textbf{0.9671} & \textbf{0.9672} & \textbf{0.9669} \\ 
        \midrule
        \multirow{2}{*}{DenseNet} 
        & FEWS & 0.9756 & 0.9762 & 0.9757 & 0.9755 \\ 
        & OEWS & \textbf{0.9901} & \textbf{0.9902} & \textbf{0.9903} & \textbf{0.9902} \\ 
        \midrule
        \multirow{2}{*}{EfficientNet} 
        & FEWS & 0.9779 & 0.9781 & 0.9782 & 0.9780 \\ 
        & OEWS & \textbf{0.9866} & \textbf{0.9867} & \textbf{0.9868} & \textbf{0.9867} \\ 
        \bottomrule
    \end{tabular}
    \end{adjustbox}
    \label{tab:FEWS_vs_OEWS}
\end{table}
The results, as summarized in Table \ref{tab:FEWS_vs_OEWS}, demonstrate that OEWS consistently outperforms FEWS on the global test dataset across all model architectures. More importantly, when evaluated on the unseen external client dataset, the model trained with OEWS achieved higher accuracy and F1 scores, indicating superior generalization capabilities. This finding is crucial as it highlights the robustness of our proposed method in handling new and diverse data, a key requirement for real-world applications in manufacturing.

Furthermore, both federated models—trained with OEWS and FEWS—showed significant improvements over centralized models in terms of accuracy and F1 scores on the global test dataset. This underscores the advantages of FL in preserving data privacy while achieving high model performance. Table \ref{tab:FEWS_vs_OEWS} summarizes the performance metrics across all CRs on the global test dataset, emphasizing the practical benefits of the OEWS strategy.

\subsection{Comparison of Federated Model (OEWS), Clients' Centralized Models, and Centralized Models on Global Test Dataset}
\begin{table}[h]
    \centering
    \caption{Performance comparison across architectures for clients' local models, the global FL model (OEWS), and the centralized model, with DenseNet noted for its efficiency and effectiveness.}
    \begin{adjustbox}{max width=\columnwidth}
    \begin{tabular}{ccccccc}
        \toprule
        \textbf{Architecture} & \textbf{Client} & \textbf{Test Acc} & \textbf{Precision} & \textbf{Recall} & \textbf{F1} \\ 
        \midrule
        \multirow{5}{*}{ResNet} 
        & Client1 & 0.2892 & 0.2497 & 0.2914 & 0.2192 \\ 
        & Client2 & 0.2660 & 0.2166 & 0.2664 & 0.2227 \\ 
        & Client3 & 0.2631 & 0.3363 & 0.2644 & 0.1734 \\ 
        & Client4 & 0.2811 & 0.2681 & 0.2832 & 0.2188 \\ 
        & FL Model & \textbf{0.9890} & \textbf{0.9890} & \textbf{0.9890} & \textbf{0.9890} \\ 
        & Centralized Model & 0.4826 & 0.5864 & 0.4845 & 0.4524 \\ 
        \midrule
        \multirow{5}{*}{VGG} 
        & Client1 & 0.4187 & 0.4746 & 0.4212 & 0.3871 \\ 
        & Client2 & 0.4686 & 0.3877 & 0.4675 & 0.4167 \\ 
        & Client3 & 0.5331 & 0.4263 & 0.5338 & 0.4730 \\ 
        & Client4 & 0.4210 & 0.3864 & 0.4242 & 0.3389 \\ 
        & FL Model & \textbf{0.9669} & \textbf{0.9671} & \textbf{0.9672} & \textbf{0.9669} \\ 
        & Centralized Model & 0.8095 & 0.8359 & 0.8101 & 0.7895 \\ 
        \midrule
        \multirow{5}{*}{DenseNet} 
        & Client1 & 0.3101 & 0.2759 & 0.3130 & 0.2675 \\ 
        & Client2 & 0.3171 & 0.2880 & 0.3180 & 0.2594 \\ 
        & Client3 & 0.3601 & 0.3286 & 0.3578 & 0.3102 \\ 
        & Client4 & 0.3089 & 0.2745 & 0.3070 & 0.2444 \\ 
        & FL Model & \textbf{0.9901} & \textbf{0.9902} & \textbf{0.9903} & \textbf{0.9902} \\ 
        & Centralized Model & 0.6440 & 0.6559 & 0.6451 & 0.6461 \\ 
        \midrule
        \multirow{5}{*}{EfficientNet} 
        & Client1 & 0.2282 & 0.2821 & 0.2302 & 0.2036 \\ 
        & Client2 & 0.2178 & 0.2379 & 0.2174 & 0.1850 \\ 
        & Client3 & 0.2346 & 0.1929 & 0.2342 & 0.2049 \\ 
        & Client4 & 0.2404 & 0.2070 & 0.2407 & 0.2107 \\ 
        & FL Model & \textbf{0.9866} & \textbf{0.9867} & \textbf{0.9868} & \textbf{0.9867} \\ 
        & Centralized Model & 0.3351 & 0.3432 & 0.3357 & 0.3271 \\ 
        \bottomrule
    \end{tabular}
    \end{adjustbox}
    \label{tab:clients_vs_fed}
\end{table}
Table \ref{tab:clients_vs_fed} compares the best local centralized model for each client against the best global federated model and the best centralized model. The FL model results in Table \ref{tab:clients_vs_fed} correspond to the OEWS model, as it was the best-performing federated model when compared to FEWS (refer to Table \ref{tab:FEWS_vs_OEWS}). In comparing the performance of the FL models to the centralized models trained on each client's dataset, we observe that the FL models outperform the centralized models across all metrics on the global test dataset. Additionally, the FL models demonstrate superior generalization when tested on the external client dataset, highlighting the robustness of the federated approach in handling unseen data. 

Moreover, DenseNet emerged as the most efficient architecture among the four, offering a balance between accuracy and computational efficiency due to its lower number of trainable parameters, which allows for faster training and superior performance in both global and external test scenarios.

\subsection{Comparison of Federated Optimal Epoch Weight Selection (OEWS) Model, Federated Final Epoch Weight Selection (FEWS), and Centralized Models on External Test Dataset}

In our evaluation, we compared the performance of centralized models and the Federated Optimal Epoch Weight Selection (OEWS) model across various architectures on an external test dataset. The results, summarized in Table \ref{tab:comparison_external_test}, clearly illustrate the advantages of the federated approach. For the ResNet architecture, the federated OEWS model achieved a substantial improvement in accuracy, reaching 0.9457 compared to the centralized model's 0.4214. Precision, recall, and F1 scores also showed significant enhancements, indicating the federated model's superior ability to generalize to new data.
\begin{table}[h]
    \centering
    \caption{Comparison of performance metrics for centralized models, federated FEWS, and federated OEWS models on external test dataset}
    \begin{adjustbox}{max width=\columnwidth}
    \begin{tabular}{ccccccc}
        \toprule
        \textbf{Model} & \textbf{Training Approach} & \textbf{Accuracy} & \textbf{Precision} & \textbf{Recall} & \textbf{F1 Score} \\ 
        \midrule
        \multirow{3}{*}{ResNet} 
        & Centralized & 0.4214 & 0.5190 & 0.4214 & 0.3511 \\ 
        & Federated FEWS & 0.9086 & 0.9234 & 0.9086 & 0.9066 \\ 
        & Federated OEWS & \textbf{0.9457} & \textbf{0.9527} & \textbf{0.9457} & \textbf{0.9458} \\ 
        \midrule
        \multirow{3}{*}{VGG} 
        & Centralized & 0.8557 & 0.8613 & 0.8557 & 0.8509 \\ 
        & Federated FEWS & 0.9100 & 0.9205 & 0.9100 & 0.9112 \\ 
        & Federated OEWS & \textbf{0.9329} & \textbf{0.9392} & \textbf{0.9329} & \textbf{0.9334} \\ 
        \midrule
        \multirow{3}{*}{DenseNet} 
        & Centralized & 0.5643 & 0.5199 & 0.5643 & 0.5297 \\ 
        & Federated FEWS & 0.9329 & 0.9429 & 0.9329 & 0.9338 \\ 
        & Federated OEWS & \textbf{0.9500} & \textbf{0.9546} & \textbf{0.9500} & \textbf{0.9501} \\ 
        \midrule
        \multirow{3}{*}{EfficientNet} 
        & Centralized & 0.2757 & 0.2748 & 0.2757 & 0.2301 \\ 
        & Federated FEWS & 0.9571 & 0.9582 & 0.9571 & 0.9573 \\ 
        & Federated OEWS & \textbf{0.9729} & \textbf{0.9733} & \textbf{0.9729} & \textbf{0.9729} \\ 
        \bottomrule
    \end{tabular}
    \end{adjustbox}
    \label{tab:comparison_external_test}
\end{table}
Similarly, the VGG model exhibited high performance under both approaches. However, the federated OEWS model maintained competitive metrics with a slightly lower accuracy (0.9329) compared to the centralized model's 0.8557 but demonstrated a balanced improvement in precision (0.9392) and F1 score (0.9334). DenseNet's federated OEWS model outperformed its centralized counterpart, achieving 0.9500 in accuracy versus 0.5643. This was mirrored in precision, recall, and F1 score improvements, emphasizing the robustness of the FL strategy. The EfficientNet architecture showed the most dramatic improvement, with the federated OEWS model attaining an accuracy of 0.9729, significantly higher than the centralized model's 0.2757. All other metrics followed this trend, underscoring the federated model's exceptional performance.

\begin{figure}[h]
    \centering
    \includegraphics[width=0.5\textwidth]{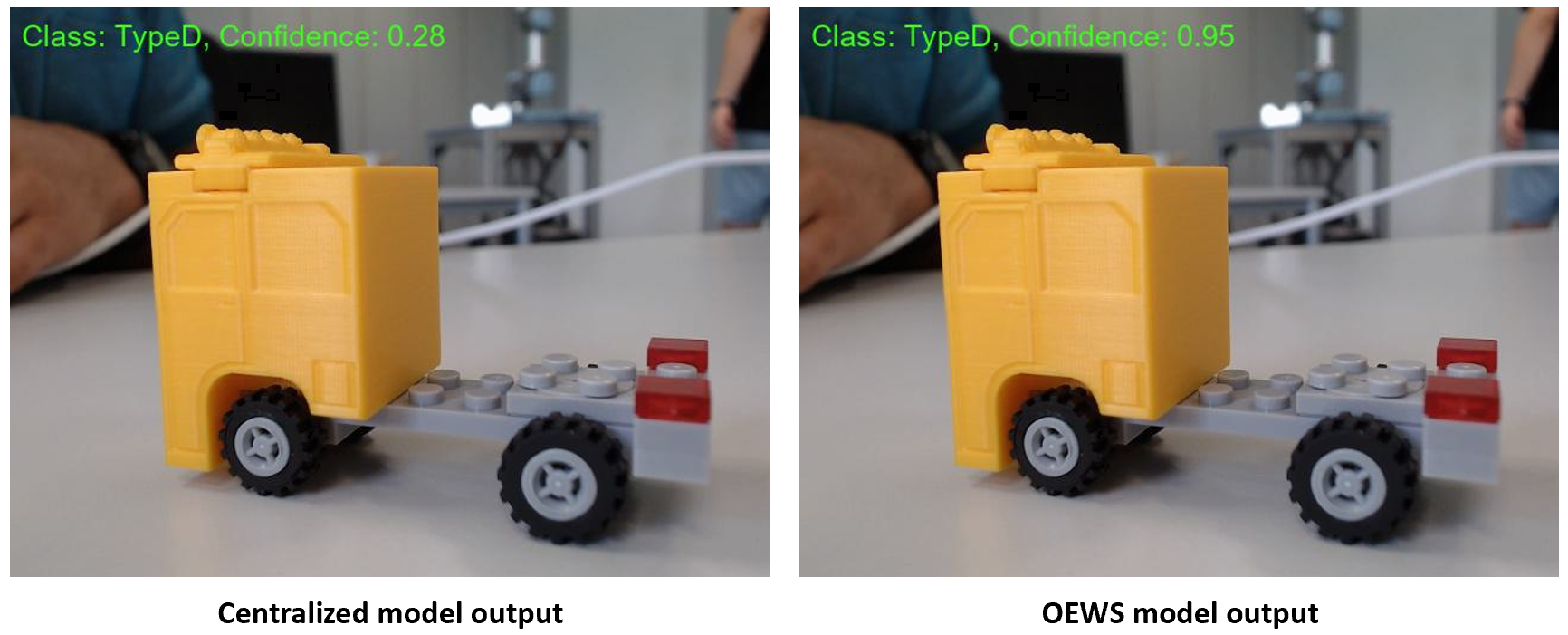}
    \caption{Comparison of confidence scores: centralized model (left) vs. federated OEWS model (right) on the external test dataset using the DenseNet architecture}
    \label{fig:comparison_confidence}
\end{figure}

From the results in Table \ref{tab:comparison_external_test}, it is evident that the Federated OEWS strategy significantly outperforms the centralized model when tested on an unseen external dataset. This is not only clear in the overall performance metrics but also in the confidence scores of correctly predicted images. For example, consider the output from the DenseNet architecture when tested on an image of a cabin with Windshield Type D. The centralized model correctly predicts the class but with a notably low confidence score of 0.28 (as shown on the left in Figure \ref{fig:comparison_confidence}). In contrast, the Federated OEWS model predicts the same class with a much higher confidence score of 0.93 (as shown on the right in Figure \ref{fig:comparison_confidence}). This demonstrates that the Federated OEWS model not only improves accuracy but also enhances the model's confidence in its predictions, which is crucial for real-world applications where reliable decision-making is required.

Overall, the federated OEWS models consistently outperformed centralized models, showcasing their enhanced capability to generalize and perform well on unseen data, which is critical for real-world applications. This comparison highlights the potential of FL, particularly in environments where collaborative data utilization is feasible and beneficial.

\section{Conclusion}
This paper examined the effectiveness of two weight selection strategies in Federated Learning (FL) within the manufacturing sector: Final Epoch Weight Selection (FEWS) and Optimal Epoch Weight Selection (OEWS). Through extensive experiments across multiple deep learning architectures and evaluations on both global and external test datasets, OEWS consistently demonstrated superior performance over FEWS. Federated models employing OEWS not only achieved higher accuracy and better generalization but also outperformed centralized models trained on individual clients' dataset. This result underscores the significant advantages of FL, particularly in preserving data privacy, maintaining data sovereignty, and delivering high performance without the need for direct access to client datasets.
The OEWS strategy, by selecting the most effective local weights for server aggregation, led to improved convergence, enhanced performance and robustness across diverse and non-IID datasets. These findings highlight the critical importance of selecting an appropriate weight aggregation strategy to maximize the benefits of FL, especially in environments characterized by heterogeneous data and a limited number of clients. This research underscores the potential of FL, particularly with OEWS, to advance collaborative machine learning in industrial settings by enhancing model robustness, efficiency, and accuracy in practical applications.

Furthermore, DenseNet121 emerged as the most efficient architecture for FL in the manufacturing context, balancing high performance with computational efficiency due to its reduced number of trainable parameters. This makes it particularly suitable for FL scenarios involving non-IID data across multiple clients.

In addition, federated models consistently outperformed traditional centralized models (trained by merging training dataset from all clients) on both global and external datasets, demonstrating a superior ability to handle unseen data combinations, diverse environments, and complex scenarios, thereby further validating the efficacy of the federated learning approach in real-world industrial applications.

\section{Limitations and Future Work}
While this study demonstrated the effectiveness of OEWS in FL within the manufacturing sector, several limitations exist. The controlled environment with a limited number of clients and datasets specific to manufacturing may restrict the generalizability of the findings to other domains with different data characteristics and larger client bases. Additionally, the impact of varying network conditions and real-world constraints such as communication overhead, resource limitations, and client device heterogeneity was not fully explored.

Future work will address these limitations by testing the proposed strategies on public datasets to assess their generalizability and scalability across different domains. Expanding the scope to include sectors such as healthcare and finance will further demonstrate the versatility of the OEWS strategy. Additionally, future research will focus on optimizing the FL process by incorporating advanced scheduling mechanisms for weight transmission and adaptive client participation, addressing challenges related to network latency, resource allocation, and dynamic client participation.
Further exploration will also be conducted to identify optimal neural network architectures for FL, particularly those with fewer trainable parameters, as demonstrated by DenseNet121 in this study. This will help better understand the trade-offs between model complexity, training efficiency, and deployment feasibility in various industrial settings.

\section*{Acknowledgment}
This work was funded by the Carl Zeiss Stiftung, Germany under the Sustainable Embedded AI project (P2021-02-009).

\bibliographystyle{IEEEtran}
\bibliography{lit}

\end{document}